\documentclass[runningheads]{llncs}
\usepackage[T1]{fontenc}
\usepackage{graphicx,verbatim}
\usepackage{booktabs}
\usepackage{multirow}
\usepackage{amsmath}
\usepackage{amssymb}
\usepackage{anyfontsize}
\usepackage{makecell}
\usepackage[hidelinks]{hyperref}
\usepackage{color}
\usepackage{makecell}
\usepackage{multirow}
\usepackage{array}
\usepackage{pifont}

\usepackage{newunicodechar}
\newunicodechar{−}{\textminus}
\usepackage{booktabs}
\usepackage{float}
\usepackage{placeins}

\begin{document}
\title{LRMIL: Efficient Low-Resolution Multiple Instance Learning via High-Resolution Knowledge Distillation for Whole Slide Image Classification}
\titlerunning{LRMIL via HR Knowledge Distillation for WSI Classification}
\begin{comment}  %% Removed for anonymized MICCAI submission
\author{First Author\inst{1}\orcidID{0000-1111-2222-3333} \and
Second Author\inst{2,3}\orcidID{1111-2222-3333-4444} \and
Third Author\inst{3}\orcidID{2222--3333-4444-5555}}
%
\authorrunning{F. Author et al.}
\institute{Princeton University, Princeton NJ 08544, USA \and
Springer Heidelberg, Tiergartenstr. 17, 69121 Heidelberg, Germany
\email{lncs@springer.com}\\
\url{http://www.springer.com/gp/computer-science/lncs} \and
ABC Institute, Rupert-Karls-University Heidelberg, Heidelberg, Germany\\
\email{\{abc,lncs\}@uni-heidelberg.de}}
\end{comment}

\author{Yonghan Shin, Won-Ki Jeong$^\dagger$}
\authorrunning{Shin et al.}
\institute{Department of Computer Science and Engineering, Korea University, Seoul, Korea \\
\email{\{dydgks592, wkjeong\}@korea.ac.kr}}

% \author{Anonymized Authors}%% Added for anonymized MICCAI submission
% \authorrunning{Anonymized Author et al.}
% \institute{Anonymized Affiliations \\
% \email{email@anonymized.com}}

\maketitle

\def\thefootnote{$\dagger$}\footnotetext{Corresponding author.}

\begin{abstract}
Multiple instance learning (MIL) has become a standard paradigm for whole slide image (WSI) analysis in digital pathology, as it enables slide-level prediction without dense annotations. Existing MIL methods typically rely on exhaustive extraction and encoding of high-resolution patches. However, this practice suffers from two critical limitations in real-world clinical settings: it struggles to capture global visual cues at lower magnifications, and incurs substantial computational overhead due to the massive number of high-resolution patches per slide.
% 
% it fails to effectively capture global visual cues that are more apparent at lower magnifications, and it incurs substantial computational overhead due to the massive number of high-resolution patches per slide, limiting scalability in real-world clinical settings.
To address these limitations, we propose an efficient low-resolution multiple instance learning (LRMIL) framework that transfers high-resolution knowledge to low-resolution representations. LRMIL adopts a two-stage distillation strategy. First, patch-level cross-resolution distillation aligns low-resolution patch embeddings with high-resolution representations. Second, slide-level knowledge distillation trains a low-resolution student MIL model under both slide-level supervision and teacher guidance. At inference time, LRMIL operates exclusively on low-resolution patches, substantially reducing data preprocessing and computational cost.
Extensive experiments on multiple WSI benchmarks demonstrate that LRMIL consistently outperforms state-of-the-art MIL methods while achieving more efficient inference. These results highlight LRMIL as a practical and scalable solution for WSI analysis in clinical pathology.
Code is available at \hyperref[]{https://github.com/hvcl/LRMIL.git}.

\keywords{Whole Slide Image  \and Multiple Instance Learning \and Knowledge Distillation.}
% Authors must provide keywords and are not allowed to remove this Keyword section.

\end{abstract}
\section{Introduction}
Whole slide image (WSI) analysis is a fundamental task in computational pathology, where learning methods must cope with extremely large images and limited annotation availability. 
Multiple instance learning (MIL) has therefore become a standard paradigm, as it enables slide-level prediction by aggregating patch-level representations without per-patch % requiring dense 
annotations. 
Recent MIL-based approaches have shown superior results in pathology applications~\cite{abmil,dsmil,clam,transmil,dtfdmil}.
However, most existing methods rely on exhaustive extraction and encoding of high-resolution (HR) patches, which introduce critical limitations in both representation capacity and computational efficiency.

Specifically, reliance on high-magnification patches alone restricts the modeling of global visual cues—such as tissue architecture and spatial context—that are more readily captured at lower magnifications.
%
%First, relying exclusively on high-magnification patches hinders the modeling of global visual cues—such as tissue architecture and spatial context—that are often more discernible at lower magnifications.
%
In addition, a single WSI may contain thousands to tens of thousands of HR patches, and the associated preprocessing overhead, including patch extraction and feature encoding, frequently dominates the overall computational cost of the pipeline.
Several recent studies attempt to alleviate this issue by filtering or selecting informative HR patches during MIL inference~\cite{hdmil,emmpd,focus,zoommil}. However, such strategies ultimately rely on HR patch encoding for final prediction and do not explicitly transfer HR semantic knowledge to low-resolution (LR) representations. As a result, LR patches alone remain insufficient for accurate inference, and the computational benefits of LR processing cannot be fully realized. Consequently, the scalability and practicality of existing HR-based MIL pipelines remain limited in real-world clinical settings.

% A natural alternative is to perform MIL using LR patches, which are substantially cheaper to process and better suited for capturing global contextual information. However, naively applying MIL at LR often results in degraded performance, as fine-grained discriminative patterns critical for accurate diagnosis may be lost. Existing multi-scale or hierarchical MIL approaches partially mitigate this issue by incorporating information from multiple magnifications, but they continue to depend on HR processing at inference time and therefore fail to eliminate the dominant preprocessing overhead.

Recently, knowledge distillation (KD) has emerged as an effective paradigm for improving efficiency while preserving performance, by transferring knowledge from a large teacher to a smaller student model~\cite{KD}. 
Originally developed for convolutional neural networks, KD has since been extended to various architectures (\textit{e.g.}, vision transformers) and has shown effectiveness across computer vision tasks~\cite{kd1,kd2,kd3,kd5,kd4}. 
In computational pathology, several recent studies have also adopted distillation-based strategies to transfer rich semantic knowledge from large models to more efficient student models~\cite{h0mini,virchow2}. 
However, these approaches mainly focus on pretraining compact or lightweight models through KD, rather than addressing the resolution-dependent inefficiencies of MIL pipelines.

To address these challenges, we propose an efficient low-resolution MIL (LRMIL) framework that decouples training resolution from inference resolution. 
LRMIL leverages fine-grained information only during training, while enabling inference to be performed exclusively on LR patches. 
Conceptually, LRMIL consists of two distillation stages:  
First, we perform patch-level cross-resolution distillation, where a frozen HR patch encoder provides supervision to train an LR encoder with the same architecture. 
This stage aims to transfer HR semantic knowledge into LR patch representations. Second, an LR-based MIL model is trained using the distilled encoder, guided by both bag-level supervision and slide-level knowledge distillation from an HR-based teacher MIL model. 
At inference time, only the student MIL model is used, operating solely on LR patches. 
This design substantially reduces data preprocessing and feature extraction costs for inference WSIs, enabling efficient and scalable deployment in clinical pathology workflows. 
%c
Our contributions can be summarized as follows:
\begin{itemize}
    \item We propose an efficient MIL framework for WSIs that is explicitly designed to perform inference using only LR patches.     
    \item We introduce a novel two-stage knowledge distillation strategy, consisting of patch-level cross-resolution distillation and slide-level distillation tailored for computational pathology.    
    \item Through extensive experiments on diverse datasets and tasks, we demonstrate that the proposed method achieves superior performance and computational efficiency compared to state-of-the-art MIL approaches.
\end{itemize}

% (1) We propose an efficient MIL framework for WSIs that is explicitly designed to perform inference using only LR patches. (2) We introduce a two-stage knowledge distillation strategy, consisting of patch-level cross-resolution distillation and slide-level distillation for MIL. 
% (3) Through extensive experiments on diverse datasets and tasks, we demonstrate that the proposed method achieves superior performance and efficiency compared to state-of-the-art MIL approaches.
\section{Methodology}

\begin{figure*}[t]
\centering
\includegraphics[width=0.99\linewidth]{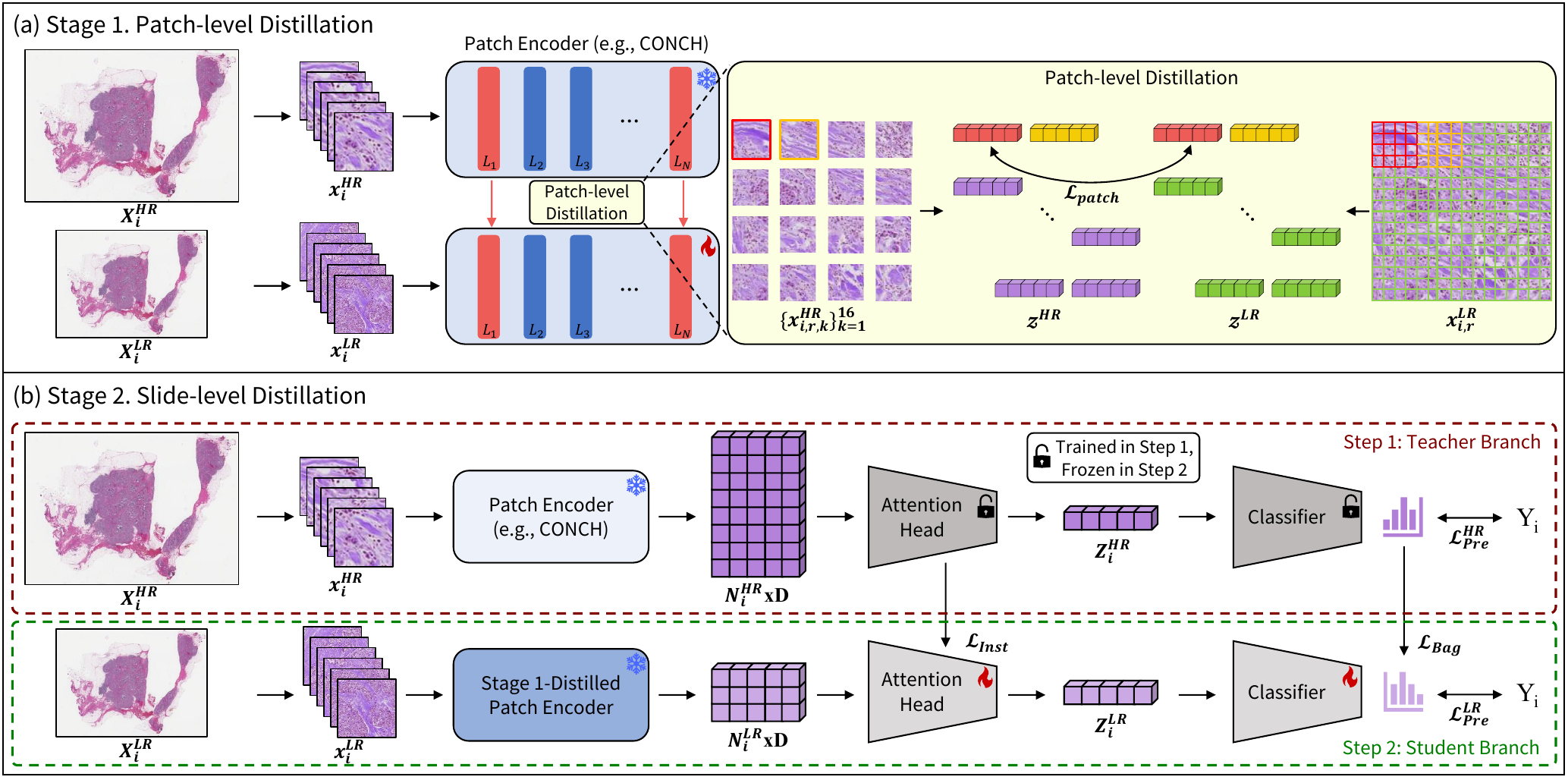}
\caption{Overview of our LRMIL framework. (a) Patch-level cross-resolution distillation. Fine-grained semantic knowledge is distilled to a coarse-level patch encoder. (b) Slide-level distillation for MIL. An LR-based student MIL model is trained using both bag-level supervision and teacher guidance.}
\label{framework}
\end{figure*}

% An overview of the proposed LRMIL is illustrated in Fig.~\ref{framework}. Conceptually, LRMIL consists of two distillation stages: (i) encoder-level cross-resolution distillation at the patch level ((a) in Fig.~\ref{framework}), and (ii) bag-level knowledge distillation for multiple instance learning((b) in Fig.~\ref{framework}). These two stages jointly enable low-resolution patches to support accurate and efficient MIL inference.

Figure~\ref{framework} illustrates the overall framework of LRMIL. 
The key idea of LRMIL is to enable accurate MIL inference using only \textit{low-resolution} patches, while transferring high-resolution knowledge during training through distillation. 
To this end, LRMIL adopts a two-stage distillation strategy consisting of patch-level cross-resolution distillation (Fig.~\ref{framework}(a)) and slide-level knowledge distillation (Fig.~\ref{framework}(b)).
% In the first stage (Fig.~\ref{framework}(a)), pre-trained high-resolution patch encoder provides supervision to train a LR encoder by aligning patch embeddings from corresponding spatial regions. This stage operates independently of MIL and aims to transfer high-resolution semantic information into low-resolution representations.
% In the second stage (Fig.~\ref{framework}(b)), a low-resolution student MIL model is trained using the distilled encoder, guided by both slide-level supervision and knowledge distillation from a high-resolution teacher MIL model.
At inference time, only the low-resolution student model is used, enabling efficient WSI analysis without high-resolution processing.

\subsection{Stage 1: Patch-level Cross-resolution Distillation}

The objective of stage~1 is to transfer fine-grained semantic knowledge from HR observations to LR representations at the patch level, independently of MIL. By aligning matched HR--LR regions, this stage encourages the LR encoder to capture both (i) structural characteristics visible at low magnification and (ii) fine-grained details present at high magnification.

Let $X_{i}^{HR}$ and $X_{i}^{LR}$ denote the $i$-th whole slide image (WSI) at high and low magnifications, respectively. We crop HR and LR patches of the same pixel size ($256 \times 256$) from $X_{i}^{HR}$ and $X_{i}^{LR}$. In our setting (e.g., $20\times$ for HR and $5\times$ for LR), one LR patch covers the same region as $K=16$ HR patches. For each LR patch $x^{LR}_{i,r} \subset X_{i}^{LR}$ covering region $r$, we therefore identify the corresponding HR patches $\{x^{HR}_{i,r,k}\}_{k=1}^{K} \subset X_{i}^{HR}$ that jointly cover the same region at higher magnification. This \emph{cross-scale matching} is used to construct distillation pairs.

We adopt a ViT-based architecture for both resolutions. 
A pre-trained HR encoder $f_{HR}(\cdot)$ is frozen and used as a teacher, while an LR encoder $f_{LR}(\cdot)$ with the same architecture is trained. 
Our distillation strategy follows prior work in the natural image domain that performs multi-layer distillation between a large teacher and a smaller student model~\cite{kd5}. 
Specifically, we adopt the same layer selection strategy (e.g., layers $L=\{0,1,11\}$ in a 12-layer ViT) and MSE-based alignment objective. 
However, instead of distilling representations from identical images, we extend this formulation to a cross-resolution setting tailored for computational pathology, where HR and LR patches correspond to the same spatial region at different magnifications.

% At each selected layer $l \in L$, we compute the CLS embedding of HR patch:
% \begin{equation}
% \mathbf{z}^{HR}_{i,r,k,l} = \mathrm{CLS}_l\big(f_{HR}(x^{HR}_{i,r,k})\big), \quad k=1,\dots,K.
% \end{equation}
% For the LR patch $x^{LR}_{i,r}$, the ViT produces a grid of image tokens. With a patch size of $16 \times 16$ pixels, a $256 \times 256$ LR patch yields a $16 \times 16$ token grid, i.e., $256$ spatial tokens. Each HR patch corresponds to a $4 \times 4$ sub-grid of LR tokens (16 tokens) covering the same physical region. Let $\mathcal{S}_{k}$ denote the index set of LR tokens matched to the $k$-th HR patch. We aggregate these matched LR tokens to form a region-level representation (denoted as \emph{Image Tokens} in (Fig.~\ref{framework}(a)):
% \begin{equation}
% \mathbf{z}^{LR}_{i,r,k,l} = \frac{1}{|\mathcal{S}_{k}|} \sum_{m \in \mathcal{S}_{k}} \mathbf{t}^{LR}_{i,r,m,l},
% \end{equation}
% where $\mathbf{t}^{LR}_{i,r,m,l}$ denotes the $m$-th LR spatial token at layer $l$ and $|\mathcal{S}_{k}|=16$ in our setting.

At each selected layer $l \in \mathcal{L}$, we extract matched HR and LR representations for each spatial region indexed by $k$. 
Specifically, for a $256 \times 256$ LR patch with a ViT patch size of $16 \times 16$ pixels, the encoder produces a $16 \times 16$ grid of spatial tokens (256 tokens in total). 
Each HR patch corresponds to a $4 \times 4$ sub-grid (16 tokens) within this LR token grid. 
Let $\mathcal{S}_{k}$ denote the index set of LR tokens matched to the $k$-th HR patch via cross-scale matching. 
We then define the HR and LR representations at layer $l$ as:
\begin{equation}
\textit{z}^{HR}_{i,r,k,l} 
= \mathrm{CLS}_l\big(f_{HR}(x^{HR}_{i,r,k})\big), \; \; \; 
\textit{z}^{LR}_{i,r,k,l} 
= \frac{1}{|\mathcal{S}_{k}|} 
\sum_{m \in \mathcal{S}_{k}} 
\mathbf{t}^{LR}_{i,r,m,l},
\quad k=1,\dots,K,
\end{equation}
where $\mathbf{t}^{LR}_{i,r,m,l}$ denotes the $m$-th LR spatial token at layer $l$, and $|\mathcal{S}_{k}| = 16$ in our setting. Here, K denotes the number of HR patches matched to a single LR patch, whereas $|\mathcal{S}_{k}|$ denotes the number of LR spatial tokens pooled for the k-th matched HR patch; although both are 16, they represent different quantities.
% This formulation explicitly aligns each HR CLS embedding with the averaged LR tokens from the spatially corresponding region.

We then perform patch-level cross-resolution distillation by aligning each HR CLS embedding with its matched LR region-level token representation. The overall stage~1 objective aggregates losses across the selected layers:
\begin{equation}
\mathcal{L}_{\mathrm{layer}}^{(l)} 
= \frac{1}{K} \sum_{k=1}^{K}
\left\|
\textit{z}^{HR}_{i,r,k,l} - \textit{z}^{LR}_{i,r,k,l}
\right\|_2^2,
\; \; \; 
\mathcal{L}_{patch}
= \sum_{l \in L} \mathcal{L}_{\mathrm{layer}}^{(l)}.
\end{equation}
% \begin{equation}
% \mathcal{L}_{stage1}
% = \sum_{l \in L} \mathcal{L}_{\mathrm{PD}}^{(l)}.
% \end{equation}

By distilling HR CLS embeddings into matched LR region-level token representations, stage~1 injects fine-grained HR semantics into spatially corresponding LR features while preserving low-magnification structural cues. 
The resulting distilled LR encoder is subsequently used to construct LR bags for MIL training in stage~2, without requiring any HR processing at inference time.

\subsection{Stage 2: Slide-level Knowledge Distillation}
\label{sec:stage2}

In stage~2, we train an attention-based MIL model that performs accurate bag-level prediction using LR patches only. 
For this, both the HR patch encoder and the distilled LR encoder from stage~1 are frozen, and only the MIL heads are trained.
% To keep the overall pipeline conceptually simple,
We optimize the teacher and student MIL models sequentially: train an HR teacher MIL model with bag-level supervision (step 1), and train an LR student MIL model with bag-level supervision and teacher guidance (step 2).

\noindent\textbf{Step 1.}
Given the $i$-th WSI $X_{i}^{HR}$, we extract and encode HR patches using the frozen HR encoder to obtain instance embeddings. 
An attention-based MIL head aggregates these embeddings into a bag-level feature $\mathbf{Z}^{HR}_i$ and produces bag-level logits $\mathbf{p}^{HR}_i$.
We train the MIL head using only the bag-level label $Y_i$ with the CE loss,
% \begin{equation}
$\mathcal{L}_{pre}^{HR} = \mathrm{CE}\!\left((\mathbf{p}^{HR}_i), \, Y_i\right)$.
% \end{equation}
 % where $\sigma(\cdot)$ denotes the softmax function. 
Since the encoder is frozen, this step is lightweight and converges quickly in practice.

\noindent\textbf{Step 2.}
We then train an LR student MIL model on $X_{i}^{LR}$ using the distilled LR encoder.
The student MIL head outputs logits $\mathbf{p}^{LR}_i$, same as step 1.
During this step, the teacher MIL model is frozen and provides guidance through both bag-level logits and instance-level attention scores.

\noindent\textit{Bag-level distillation.}
We distill the teacher's bag-level prediction by minimizing the KL divergence between the softened distributions:
\begin{equation}
\mathcal{L}_{\mathrm{bag}} = 
\mathrm{KL}\!\left(
\mathtt{softmax}(\mathbf{p}^{HR}_i / T_1) \,\Vert\, \mathtt{softmax}(\mathbf{p}^{LR}_i / T_1)
\right), 
% \; \; \text{where} \; T \; \text{is temperature.}
\end{equation}
where $T_1$ is the temperature.

\noindent\textit{Instance-level distillation.}
While bag-level distillation aligns slide-level predictions, it does not explicitly enforce patch-level consistency. 
Therefore, we introduce instance-level distillation to further align attention patterns across resolutions.
Let $\boldsymbol{\alpha}^{HR}_i$ and $\boldsymbol{\alpha}^{LR}_i$ denote the attention weights produced by the teacher and student MIL heads, respectively. 
To align attention across resolutions, we leverage the cross-scale matching defined in stage~1, where each LR patch corresponds to $K{=}16$ HR patches covering the same spatial region r. 
For each LR patch, we construct a region-level teacher attention score, denoted as $\tilde{\boldsymbol{\alpha}}^{HR}_{i,r}$, by averaging the teacher attention weights of the matched HR patches. 
This averaged score is then used to supervise the corresponding LR patch attention.

\noindent(i) Soft attention matching: 
We match the teacher and student attention distributions using KL divergence on softmaxed scores:
\begin{equation}
\mathcal{L}_{\mathrm{inst}}^{\mathrm{soft}} =
\mathrm{KL}\!\left(
\mathtt{softmax}(\tilde{\boldsymbol{\alpha}}^{HR}_i / T_2)
\,\Vert\,
\mathtt{softmax}(\boldsymbol{\alpha}^{LR}_i / T_2)
\right), 
% \text{where} \; \tilde{\alpha}^{HR}_i \; \text{stacks} \;\tilde{\alpha}^{HR}_{i,r} / \;\text{over LR patches.}
\end{equation}
where $\tilde{\boldsymbol{\alpha}}^{HR}_i$ stacks $\tilde{\boldsymbol{\alpha}}^{HR}_{i,r}$ over LR patches, and $T_2$ is the temperature.

\noindent(ii) Hard top-$k$/bottom-$k$ supervision: 
To further encourage discrimination, we treat the regions with the highest and lowest teacher scores as pseudo-positive and pseudo-negative instances, inspired by CLAM~\cite{clam}.
Let $\mathcal{P}_i$ and $\mathcal{N}_i$ denote the indices of the top-$k$ and bottom-$k$ elements of $\tilde{\boldsymbol{\alpha}}^{HR}_i$, respectively.
We define a binary target $t_{i,r}{=}1$ for $r\in\mathcal{P}_i$ and $t_{i,r}{=}0$ for $r\in\mathcal{N}_i$, and optimize:
\begin{equation}
\mathcal{L}_{\mathrm{inst}}^{\mathrm{hard}} =
\frac{1}{|\mathcal{P}_i|+|\mathcal{N}_i|}
\sum_{r \in \mathcal{P}_i \cup \mathcal{N}_i}
\mathrm{BCE}\!\left(\boldsymbol{\alpha}^{LR}_{i,r}, \, t_{i,r}\right),
\end{equation}
% where $k$ is a hyperparameter.

\noindent\textit{Overall objective.}
The student MIL model is trained with a weighted sum of bag-level supervision and distillation losses:
\begin{equation}
\mathcal{L}_{\mathrm{slide}} =
\mathcal{L}_{pre}^{LR}
+ \lambda_{\mathrm{bag}} \mathcal{L}_{\mathrm{bag}}
+ \lambda_{\mathrm{soft}} \mathcal{L}_{\mathrm{inst}}^{\mathrm{soft}}
+ \lambda_{\mathrm{hard}} \mathcal{L}_{\mathrm{inst}}^{\mathrm{hard}},
\end{equation}
where $\mathcal{L}_{pre}^{LR}=\mathrm{CE}((\mathbf{p}^{LR}_i),Y_i)$ and $\lambda_{\mathrm{bag}},\lambda_{\mathrm{soft}}, \text{and }\lambda_{\mathrm{hard}}$ control contributions of each loss term.
At inference time, we discard the HR teacher branch entirely and use only the LR student MIL model, enabling efficient WSI analysis without HR patch extraction or encoding.

\section{Experiments and Results}
\begin{table*}[t]
\centering
\caption{Subtype classification results on four datasets. All results except BRCA\textsuperscript{*} correspond to histologic subtype classification. The best result is marked in \textbf{bold}.}
\fontsize{8pt}{10pt}\selectfont
% \resizebox{\textwidth}{!}{
{
\begin{tabular}{c l|ccc|ccc|ccc|ccc|ccc}
\Xhline{2\arrayrulewidth}
\multirow{2}{*}{} & \multirow{2}{*}{Method}
& \multicolumn{3}{c|}{TCGA-BRCA} 
& \multicolumn{3}{c|}{TCGA-NSCLC} 
& \multicolumn{3}{c|}{TCGA-RCC} 
& \multicolumn{3}{c|}{BRACS} 
& \multicolumn{3}{c}{BRCA\textsuperscript{*}} \\
\cline{3-5} \cline{6-8} \cline{9-11}
\cline{12-14} \cline{15-17}
& & Acc & AUC & $t$ 
& Acc & AUC & $t$ 
& Acc & AUC & $t$ 
& Acc & AUC & $t$ 
& Acc & AUC & $t$ \\
\hline

\multirow{8}{*}{\rotatebox{90}{\footnotesize Full HR}} 
& Max-P  & 85.2 & 86.5 & 49  & 75.8 & 84.1 & 121 & 91.7 & 97.5 & 134 & 50.6 & 82.8 & 27 & 60.1 & 78.0 & 49 \\
& Mean-P & 88.5 & 91.6 & 49  & 79.4 & 89.1 & 121 & 92.6 & 97.8 & 134 & 51.1 & 80.7 & 27 & 64.3 & 80.0 & 49 \\
& ABMIL        & 88.3 & 91.1 & 49  & 81.0 & 89.0 & 121 & 92.0 & 97.5 & 134 & 59.5 & 83.4 & 27 & 62.3 & 81.7 & 49 \\
&CLAM-MB      & 86.9 & 91.6 & 49  & 81.9 & 91.5 & 121 & 94.0 & 98.1 & 134 & \textbf{60.3} & 83.3 & 27 & 63.5 & 81.9 & 49 \\
&CLAM-SB      & 89.6 & 91.8 & 49  & 80.7 & 89.5 & 121 & 93.1 & 97.0 & 134 & 55.3 & 82.9 & 27 & 64.0 & 82.0 & 49 \\
&DSMIL        & 87.5 & 91.1 & 49  & 82.2 & 90.8 & 121 & 93.8 & 98.7 & 134 & 58.8 & \textbf{83.7} & 27 & \textbf{66.8} & 82.6 & 49 \\
&TransMIL     & 88.2 & 92.1 & 49  & 81.1 & 89.6 & 121 & 92.9 & 98.1 & 135 & 52.9 & 83.6 & 27 & 61.6 & 80.5 & 49 \\
&DTFD-MIL    & 88.5 & 91.1 & 49  & 83.1 & \textbf{92.1} & 121 & 93.8 & 97.9 & 134 & 57.3 & 82.9 & 27 & 64.3 & 82.7 & 49 \\
\hline
\multirow{2}{*}{\rotatebox{90}{\footnotesize SC}}
&ZOOMMIL     & 88.6 & 92.2 & 9  & 82.4 & 91.7 & 22  & 93.7 & 98.2 & 24  & 57.9 & 82.5 & 5 & 66.2 & \textbf{82.8} & 8 \\
&HDMIL        & 89.8 & 91.9 & 46  & 83.1 & 91.0 & 117 & 93.1 & 98.5 & 133 & 59.8 & 82.9 & 25 & 65.8 & 81.8 & 45 \\
\hline
& \textbf{Ours} & \textbf{90.7} & \textbf{92.3} & \textbf{4} 
             & \textbf{83.2} & 90.6 & \textbf{10}
             & \textbf{94.2} & \textbf{99.0} & \textbf{11}
             & 58.8 & \textbf{83.7} & \textbf{3}
             & \textbf{66.8} & \textbf{82.8} & \textbf{4} \\
\Xhline{2\arrayrulewidth}
\end{tabular}
}
\label{cls}
\end{table*}

\begin{table*}[t]
\centering
\caption{Survival prediction results on five datasets. The best result is marked in \textbf{bold}.}
\fontsize{8pt}{10pt}\selectfont
% \resizebox{\textwidth}{!}{
{
\begin{tabular}{c l|ccc|ccc|ccc|ccc|ccc}
\Xhline{2\arrayrulewidth}
\multirow{2}{*}{} & \multirow{2}{*}{Method}
& \multicolumn{3}{c|}{TCGA-BRCA} 
& \multicolumn{3}{c|}{TCGA-LUAD} 
& \multicolumn{3}{c|}{TCGA-LUSC} 
& \multicolumn{3}{c|}{TCGA-KIRP} 
& \multicolumn{3}{c}{TCGA-KIRC} \\
\cline{3-5} \cline{6-8} \cline{9-11}
\cline{12-14} \cline{15-17}
& & Acc & AUC & $t$ 
& Acc & AUC & $t$
& Acc & AUC & $t$
& Acc & AUC & $t$
& Acc & AUC & $t$ \\
\hline

\multirow{8}{*}{\rotatebox{90}{\footnotesize Full HR}} 
&Max-P  & 85.1 & 66.7 & 49  & 62.2 & 60.9 & 62 & 56.4 & 62.5 & 59 & 82.8 & 51.5 & 45  & 58.7 & 71.0 & 80 \\
&Mean-P & 85.7 & 67.8 & 49  & 61.6 & 60.8 & 62 & 57.3 & 63.1 & 59 & 82.9 & 62.7 & 45  & 68.4 & 70.5 & 80 \\
&ABMIL       & 86.2 & 66.5 & 49  & 62.7 & 61.3 & 62 & 58.4 & 61.6 & 59 & 83.1 & 61.6 & 45  & 68.2 & 69.2 & 80 \\
&CLAM-MB     & 85.4 & 67.3 & 49  & 62.9 & 59.6 & 62 & 60.0 & 63.3 & 59 & 83.1 & 60.5 & 45  & 68.3 & 71.7 & 80 \\
&CLAM-SB     & 86.4 & 67.4 & 49  & 61.9 & 62.0 & 62 & 57.5 & 62.5 & 59 & 79.2 & 61.2 & 45  & 68.4 & 71.2 & 80 \\
&DSMIL       & 84.7 & 65.9 & 49  & 61.4 & 61.3 & 62 & 59.2 & 63.4 & 59 & 82.0 & 59.7 & 45  & 68.0 & 70.4 & 80 \\
&TransMIL     & 86.2 & 64.7 & 49  & 57.8 & 54.7 & 62 & 53.9 & 56.1 & 59 & 79.5 & 55.7 & 45  & 66.6 & 66.3 & 80 \\
&DTFD-MIL    & 84.0 & 64.2 & 49  & 62.4 & 61.3 & 62 & 60.6 & 62.3 & 59 & 80.6 & 61.0 & 45  & 68.9 & \textbf{73.9} & 80 \\
\hline
\multirow{2}{*}{\rotatebox{90}{\footnotesize SC}} 
&ZOOMMIL     & 86.7 & 65.3 & 12  & 60.5 & 59.4 & 12  & 58.3 & 62.4 & 11  & 82.4 & \textbf{63.0} & 9  & 68.9 & 71.3 & 14 \\
&HDMIL        & 84.6 & 66.3 & 44  & 62.1 & 62.0 & 60 & 59.9 & 63.4 & 55 & 83.0 & 62.7 & 40  & 68.4 & 72.7 & 74 \\
\hline
&\textbf{Ours} & \textbf{90.0} & \textbf{92.0} & \textbf{4} 
             & \textbf{63.0} & \textbf{63.3} & \textbf{5}
             & \textbf{61.7} & \textbf{63.5} & \textbf{5}
             & \textbf{83.3} & 60.0 & \textbf{4}
             & \textbf{70.9} & 71.6 & \textbf{7} \\
\Xhline{2\arrayrulewidth}
\end{tabular}
}
\label{sur}
\end{table*}

\subsection{Experimental Setup}
% \subsubsection{Datasets and evaluation metrics.}
\noindent\textbf{Datasets and Evaluation Metrics.}
We evaluate LRMIL on three downstream tasks: histologic and molecular subtype classification, and survival prediction. 
For histologic classification, we use four public datasets: TCGA-BRCA (IDC vs.\ ILC), TCGA-NSCLC (LUAD vs.\ LUSC), TCGA-RCC (KIRP vs.\ KIRC vs.\ KICH), and BRACS (7 classes)~\cite{bracs,tcga}. 
For molecular classification, we use TCGA-BRCA to predict LumA, LumB, Basal, and Her2. 
For survival prediction, we use TCGA cohorts (BRCA, LUAD, LUSC, KIRP, and KIRC) and formulate the task as binary classification (alive vs.\ dead) based on overall survival status. 
Note that BRCA\textsuperscript{*} means molecular subtype classification on TCGA-BRCA.
We report Accuracy (Acc) and Area Under the ROC Curve (AUC) for all tasks.  
To evaluate computational efficiency, we measure total inference time $t$ (patch extraction, feature encoding, and MIL aggregation), reported in ($\times 10^{4}$ s).

% \subsubsection{Implementation details.}
\noindent\textbf{Implementation Details.}
We compare the proposed method with three groups of baselines: 
% (1) conventional pooling methods (max- and mean-pooling), 
(1) conventional pooling methods including max- and mean-pooling (Max-P and Mean-P),
(2) state-of-the-art MIL including ABMIL~\cite{abmil}, CLAM-MB, CLAM-SB~\cite{clam}, DSMIL~\cite{dsmil}, TransMIL~\cite{transmil}, and DTFD-MIL~\cite{dtfdmil}, and 
(3) efficiency-oriented methods that reduce preprocessing by selectively cropping informative regions at inference time, including ZOOMMIL~\cite{zoommil} and HDMIL~\cite{hdmil}. 
For fair comparison, all methods, including HR and LR encoders in LRMIL, use the same CONCH~\cite{conch} vision encoder as the feature extractor. 
All models are trained using the Adam optimizer at a learning rate of $1\times10^{-4}$, early stopping with a patience of 20 epochs, and 5-fold cross-validation. 
For the slide-level distillation, we set the loss weights to
\(\lambda_{\mathrm{bag}} = 0.5\),
\(\lambda_{\mathrm{soft}} = 0.5\), and
\(\lambda_{\mathrm{hard}} = 0.3\).

\begin{figure}[t]
\centering
\includegraphics[width=0.90\linewidth]{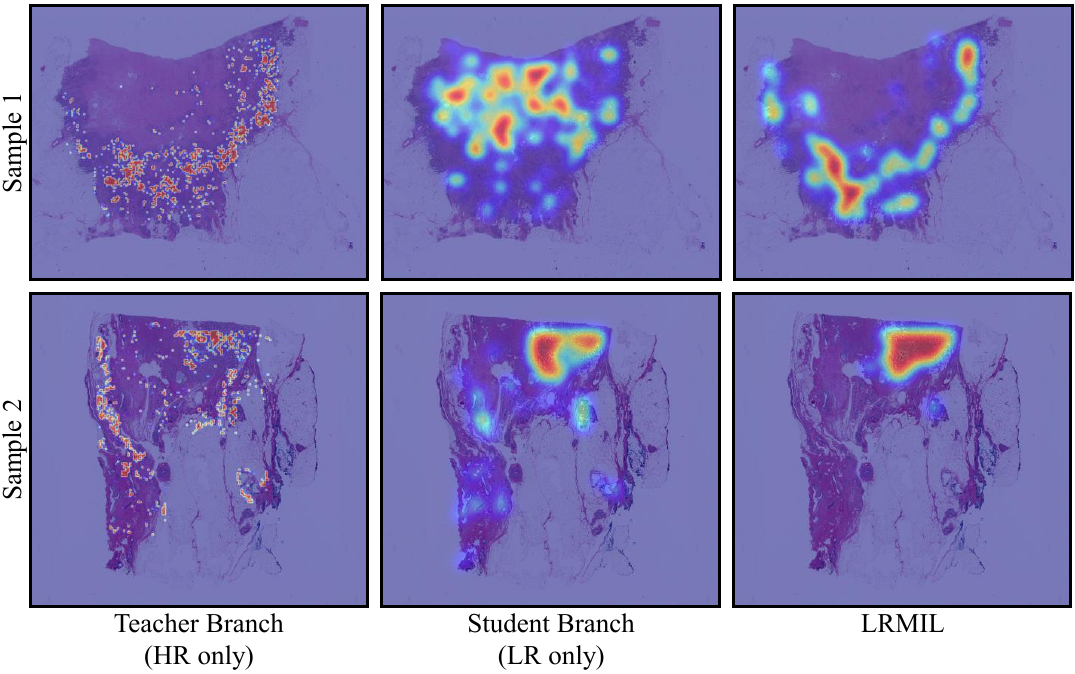}
\caption{Visual comparison of attention heatmaps.}
\label{heatmap}
\end{figure}

\subsection{Comparison Results}
\noindent\textbf{Quantitative Results.}
Tables~\ref{cls} and~\ref{sur} show the results of subtype classification and survival prediction, respectively. 
“Full HR” uses all HR patches, while “SC” (selective cropping) selects a subset of patches during inference for efficiency.
Compared to other methods, LRMIL demonstrates strong performance across all datasets and downstream tasks. 
Notably, by using exclusively on LR patches, LRMIL reduces total inference time by \textit{more than an order of magnitude} compared to HR-based methods. 
Furthermore, even compared to efficient “SC” methods, LRMIL achieves both faster inference and superior performance.

\noindent\textbf{Qualitative Results.}
Figure~\ref{heatmap} presents representative attention heatmaps, comparing the HR teacher, LR-only student, and LRMIL. 
In the first case, the teacher focuses on tumor regions while the LR-only student fails to localize them accurately; in the second case, the LR-only student captures tumor areas that the teacher overlooks. 
In contrast, LRMIL consistently highlights clinically relevant tumor regions in both cases, suggesting that the model can integrate fine-grained details and global contextual cues across magnifications effectively.

\begin{table*}[t]
\centering
\caption{Ablation study of each component in LRMIL. Each dataset corresponds to a different task: histologic subtype classification, molecular subtype classification, and survival prediction. The best result is marked in \textbf{bold}.}
\setlength{\tabcolsep}{2pt}
\fontsize{8pt}{10pt}\selectfont
\begin{tabular}{cccc|cc|cc|cc}
\Xhline{2\arrayrulewidth}
\multicolumn{4}{c|}{Loss Function} 
& \multicolumn{2}{c|}{TCGA-RCC} 
& \multicolumn{2}{c|}{BRCA\textsuperscript{*}}
& \multicolumn{2}{c}{TCGA-LUSC} \\
\cline{1-4} \cline{5-6} \cline{7-8}
\cline{9-10}
$\mathcal{L}_{patch}$ 
& $\mathcal{L}_{bag}$ 
& $\mathcal{L}_{inst}^{soft}$ 
& $\mathcal{L}_{inst}^{hard}$ 
& Acc & AUC 
& Acc & AUC
& Acc & AUC \\
\hline
&  &  &   & 91.4 & 97.8 & 62.8 & 79.8 & 57.4 & 62.0 \\
\checkmark&  &  &    & 92.6 & 98.0 & 63.1 & 80.7 & 59.1 & 62.4 \\
\checkmark& \checkmark &  &    & 92.8 & 97.8 & 65.4 & 80.9 & 58.0 & \textbf{63.7}  \\
\checkmark& \checkmark & \checkmark &    & \textbf{94.2} & 98.7 & 66.2 & 82.1 & 61.6 & 62.8  \\
\checkmark& \checkmark &  & \checkmark   & 93.8 & 98.5 & 65.6 & 81.2 & 61.4 & 63.0  \\
\checkmark & \checkmark & \checkmark & \checkmark 
& \textbf{94.2} & \textbf{99.0} 
& \textbf{67.0} & \textbf{82.9} & \textbf{61.7} & 63.5 \\
\Xhline{2\arrayrulewidth}
\label{ablation}
\end{tabular}
\label{ablation}
\end{table*}

\subsection{Ablation Study}
 \noindent\textbf{Component Ablation.}
% We analyze the contribution of distillation terms ($\mathcal{L}_{\mathrm{Bag}}$, $\mathcal{L}_{\mathrm{Inst}}^{soft}$ and $\mathcal{L}_{\mathrm{Inst}}^{hard}$)
% on representative datasets from each task. 
% As shown in Table~\ref{ablation}, bag-level distillation improves performance over the baseline, and incorporating instance-level supervision further boosts results. 
% The combination of all three losses consistently achieves the best performance, demonstrating that instance-level alignment provides complementary benefits beyond bag-level matching.
We analyze the contribution of loss terms ($\mathcal{L}_{\mathrm{patch}}$ in stage 1, and $\mathcal{L}_{\mathrm{bag}}$, $\mathcal{L}_{\mathrm{inst}}^{soft}$ and $\mathcal{L}_{\mathrm{inst}}^{hard}$ in stage 2)
on representative datasets from each task in Table~\ref{ablation}.
The comparison between the first and second rows, both without MIL-level distillation, indicates that stage 1 successfully distills essential diagnostic information into a compact representation.
Bag-level distillation improves performance over the baseline, and incorporating instance-level supervision further boosts results.
Notably, adding $\mathcal{L}_{\mathrm{inst}}^{soft}$ leads to a substantial improvement, underscoring the effectiveness of soft instance-level attention matching.
The combination of all three losses achieves the best performance, demonstrating that instance-level alignment provides complementary benefits beyond bag-level matching.

\noindent\textbf{Sensitivity to Top-$k$.}
We vary $k \in \{2,4,8,16,32,64\}$ and evaluate performance on subtype classification and survival prediction, respectively. 
The performance remains relatively stable across different values of $k$, without a clear monotonic trend. 
This suggests that the proposed hard supervision is robust to the choice of $k$. 
In our experiments, we select $k=4$ as it achieves the best overall performance.

% \begin{figure}[t]
% \centering
% % \caption{Ablation study of each component (up) and hyperparameter k in stage 2 (down).}
% % \input{tables/table_ablation}
% \includegraphics[width=0.76\linewidth]{figures/ablation_plot_가로.pdf}
% \caption{Ablation study of hyperparameter $k$ in instance-level hard supervision.}
% \label{ablation}
% \end{figure}
\section{Conclusion}
In this work, we proposed LRMIL, a low-resolution multiple instance learning framework for efficient whole slide image analysis. 
By introducing patch-level cross-resolution distillation and slide-level knowledge distillation, LRMIL effectively transfers fine-grained high-resolution semantics to low-resolution representations. 
Extensive experiments demonstrate that LRMIL achieves superior or competitive performance while substantially reducing inference time. 
These results highlight the potential of leveraging cross-resolution distillation to enable efficient and scalable WSI analysis in real-world clinical settings. 
For future work, we plan to extend our framework to more general settings (e.g., regression).

% our method focusing on classification task, we plan to extend 

% While the current study focuses on classification tasks, extending the proposed framework to more general settings (e.g., regression) remains an important direction for future work.

\begin{credits}
\subsubsection{\ackname}
This work was supported in part by the National Research Foundation of Korea under Grant RS-2024-00349697 and Grant RS-2021-NR060143; in part by the Institute for Information and Communications Technology Planning and Evaluation under Grant IITP-2026-RS-2020-II201819; in part by the Technology Development Program funded by the Ministry of SMEs and Startups (MSS), South Korea, under Grant RS-2024-00437796; in part by the National Research Council of Science and Technology (NST) grant funded by Korean Government [Ministry of Science and Information and Communications Technology (MSIT)] under Grant GTL24031-000; and in part by Korea University Grant.

\subsubsection{\discintname}
The authors have no competing interests to declare that are relevant to the
content of this article.
\end{credits}

\bibliographystyle{splncs04}
\bibliography{template}




% LNCS Homepage, \url{http://www.springer.com/lncs}, last accessed 2023/10/25
% \end{thebibliography}
\end{document}